# Learning to Segment Object Candidates

**Pedro O. Pinheiro**[*]
pedro@opinheiro.com

**Ronan Collobert**
locronan@fb.com

**Piotr Dollár**
pdollar@fb.com

Facebook AI Research

## Abstract

Recent object detection systems rely on two critical steps: (1) a set of object proposals is predicted as efficiently as possible, and (2) this set of candidate proposals is then passed to an object classifier. Such approaches have been shown they can be fast, while achieving the state of the art in detection performance. In this paper, we propose a new way to generate object proposals, introducing an approach based on a discriminative convolutional network. Our model is trained jointly with two objectives: given an image patch, the first part of the system outputs a class-agnostic segmentation mask, while the second part of the system outputs the likelihood of the patch being centered on a full object. At test time, the model is efficiently applied on the whole test image and generates a set of segmentation masks, each of them being assigned with a corresponding object likelihood score. We show that our model yields significant improvements over state-of-the-art object proposal algorithms. In particular, compared to previous approaches, our model obtains substantially higher object recall using fewer proposals. We also show that our model is able to generalize to unseen categories it has not seen during training. Unlike all previous approaches for generating object masks, we do not rely on edges, superpixels, or any other form of low-level segmentation.

## 1 Introduction

Object detection is one of the most foundational tasks in computer vision [21]. Until recently, the dominant paradigm in object detection was the sliding window framework: a classifier is applied at every object location and scale [4, 8, 32]. More recently, Girshick *et al.* [10] proposed a two-phase approach. First, a rich set of object proposals (*i.e.*, a set of image regions which are likely to contain an object) is generated using a fast (but possibly imprecise) algorithm. Second, a convolutional neural network classifier is applied on each of the proposals. This approach provides a notable gain in object detection accuracy compared to classic sliding window approaches. Since then, most state-of-the-art object detectors for both the PASCAL VOC [7] and ImageNet [5] datasets rely on object proposals as a first preprocessing step [10, 15, 33].

Object proposal algorithms aim to find diverse regions in an image which are likely to contain objects. For efficiency and detection performance reasons, an ideal proposal method should possess three key characteristics: (i) high recall (*i.e.*, the proposed regions should contain the maximum number of possible objects), (ii) the high recall should be achieved with the minimum number of regions possible, and (iii) the proposed regions should match the objects as accurately as possible.

In this paper, we present an object proposal algorithm based on Convolutional Networks (ConvNets) [20] that satisfies these constraints better than existing approaches. ConvNets are an important class of algorithms which have been shown to be state of the art in many large scale object recognition tasks. They can be seen as a hierarchy of trainable filters, interleaved with non-linearities and pooling. ConvNets saw a resurgence after Krizhevsky *et al.* [18] demonstrated that they perform very well on the ImageNet classification benchmark. Moreover, these models learn sufficiently general image features, which can be transferred to many different tasks [10, 11, 3, 22, 23].

---

[*]Pedro O. Pinheiro is with the Idiap Research Institute in Martigny, Switzerland and Ecole Polytechnique Fédérale de Lausanne (EPFL) in Lausanne, Switzerland. This work was done during an internship at FAIR.

Given an input image patch, our algorithm generates a class-agnostic mask and an associated score which estimates the likelihood of the patch fully containing a centered object (without any notion of an object category). The core of our model is a ConvNet which jointly predicts the mask and the object score. A large part of the network is shared between those two tasks: only the last few network layers are specialized for separately outputting a mask and score prediction. The model is trained by optimizing a cost function that targets both tasks simultaneously. We train on MS COCO [21] and evaluate the model on two object detection datasets, PASCAL VOC [7] and MS COCO.

By leveraging powerful ConvNet feature representations trained on ImageNet and adapted on the large amount of segmented training data available in COCO, we are able to beat the state of the art in object proposals generation under multiple scenarios. Our most notable achievement is that our approach beats other methods by a large margin while considering a smaller number of proposals. Moreover, we demonstrate the generalization capabilities of our model by testing it on object categories not seen during training. Finally, unlike all previous approaches for generating segmentation proposals, we do not rely on edges, superpixels, or any other form of low-level segmentation. Our approach is the first to learn to generate segmentation proposals directly from raw image data.

The paper is organized as follows: §2 presents related work, §3 describes our architecture choices, and §4 describes our experiments in different datasets. We conclude in §5.

## 2 Related Work

In recent years, ConvNets have been widely used in the context of object recognition. Notable systems are AlexNet [18] and more recently GoogLeNet [29] and VGG [27], which perform exceptionally well on ImageNet. In the setting of object detection, Girshick *et al.* [10] proposed R-CNN, a ConvNet-based model that beats by a large margin models relying on hand-designed features. Their approach can be divided into two steps: selection of a set of salient object proposals [31], followed by a ConvNet classifier [18, 27]. Currently, most state-of-the-art object detection approaches [30, 12, 9, 25] rely on this pipeline. Although they are slightly different in the classification step, they all share the first step, which consist of choosing a rich set of object proposals.

Most object proposal approaches leverage low-level grouping and saliency cues. These approaches usually fall into three categories: (1) objectness scoring [1, 34], in which proposals are extracted by measuring the objectness score of bounding boxes, (2) seed segmentation [14, 16, 17], where models start with multiple seed regions and generate separate foreground-background segmentation for each seed, and (3) superpixel merging [31, 24], where multiple over-segmentations are merged according to various heuristics. These models vary in terms of the type of proposal generated (bounding boxes or segmentation masks) and if the proposals are ranked or not. For a more complete survey of object proposal methods, we recommend the recent survey from Hosang *et al.* [13].

Although our model shares high level similarities with these approaches (we generate a set of ranked segmentation proposals), these results are achieved quite differently. All previous approaches for generating segmentation masks, including [17] which has a learning component, rely on low-level segmentations such as superpixels or edges. Instead, we propose a data-driven discriminative approach based on a deep-network architecture to obtain our segmentation proposals.

Most closely related to our approach, *Multibox* [6, 30] proposed to train a ConvNet model to generate bounding box object proposals. Their approach, similar to ours, generates a set of ranked class-agnostic proposals. However, our model generates segmentation proposals instead of the less informative bounding box proposals. Moreover, the model architectures, training scheme, etc., are quite different between our approach and [30]. More recently, *Deepbox* [19] proposed a ConvNet model that learns to rerank proposals generated by *EdgeBox*, a bottom-up method for bounding box proposals. This system shares some similarities to our scoring network. Our model, however, is able to generate the proposals and rank them in one shot from the test image, directly from the pixel space. Finally, concurrently with this work, Ren et al. [25] proposed 'region proposal networks' for generating box proposals that shares similarities with our work. We emphasize, however, that unlike all these approaches our method generates segmentation masks instead of bounding boxes.

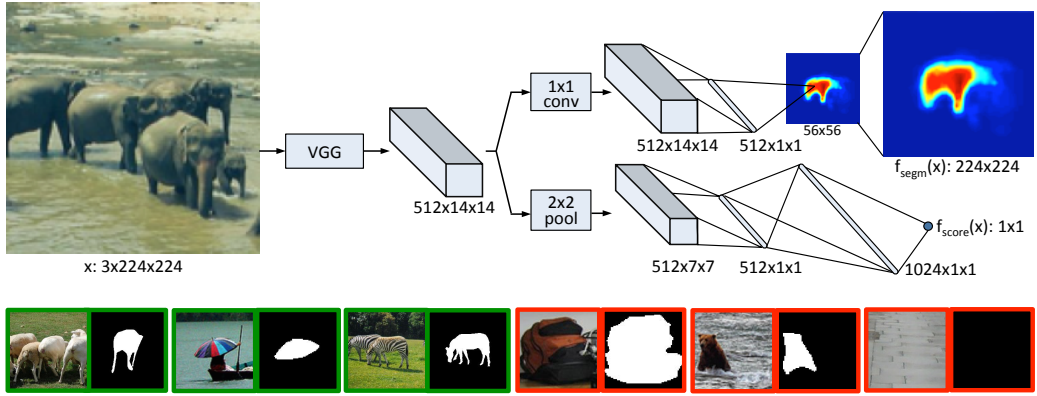

Figure 1: **(Top)** Model architecture: the network is split into two branches after the shared feature extraction layers. The top branch predicts a segmentation mask for the the object located at the center while the bottom branch predicts an object score for the input patch. **(Bottom)** Examples of training triplets: input patch $x$, mask $m$ and label $y$. Green patches contain objects that satisfy the specified constraints and therefore are assigned the label $y = 1$. Note that masks for negative examples (shown in red) are not used and are shown for illustrative purposes only.

## 3 DeepMask Proposals

Our object proposal method predicts a segmentation mask given an input patch, and assigns a score corresponding to how likely the patch is to contain an object.

Both mask and score predictions are achieved with a single convolutional network. ConvNets are flexible models which can be applied to various computer vision tasks and they alleviate the need for manually designed features. Their flexible nature allows us to design a model in which the two tasks (mask and score predictions) can share most of the layers of the network. Only the last layers are task-specific (see Figure 1). During training, the two tasks are learned jointly. Compared to a model which would have two distinct networks for the two tasks, this architecture choice reduces the capacity of the model and increases the speed of full scene inference at test time.

Each sample $k$ in the training set is a triplet containing (1) the RGB input patch $x_k$, (2) the binary mask corresponding to the input patch $m_k$ (with $m_k^{ij} \in \{\pm 1\}$, where $(i, j)$ corresponds to a pixel location on the input patch) and (3) a label $y_k \in \{\pm 1\}$ which specifies whether the patch contains an object. Specifically, a patch $x_k$ is given label $y_k = 1$ if it satisfies the following constraints:

  (i) the patch contains an object roughly centered in the input patch
  (ii) the object is fully contained in the patch and in a given scale range

Otherwise, $y_k = -1$, even if an object is partially present. The positional and scale tolerance used in our experiments are given shortly. Assuming $y_k = 1$, the ground truth mask $m_k$ has positive values only for the pixels that are part of the *single* object located in the center of the patch. If $y_k = -1$ the mask is not used. Figure 1, bottom, shows examples of training triplets.

Figure 1, top, illustrates an overall view of our model, which we call *DeepMask*. The top branch is responsible for predicting a high quality object segmentation mask and the bottom branch predicts the likelihood that an object is present and satisfies the above two constraints. We next describe in detail each part of the architecture, the training procedure, and the fast inference procedure.

### 3.1 Network Architecture

The parameters for the layers shared between the mask prediction and the object score prediction are initialized with a network that was pre-trained to perform classification on the ImageNet dataset [5]. This model is then fine-tuned for generating object proposals during training. We choose the VGG-A architecture [27] which consists of eight $3 \times 3$ convolutional layers (followed by ReLU non-linearities) and five $2 \times 2$ max-pooling layers and has shown excellent performance.

As we are interested in inferring segmentation masks, the spatial information provided in the convolutional feature maps is important. We therefore remove all the final fully connected layers of the VGG-A model. Additionally we also discard the last max-pooling layer. The output of the shared layers has a downsampling factor of 16 due to the remaining four $2 \times 2$ max-pooling layers; given an input image of dimension $3 \times h \times w$, the output is a feature map of dimensions $512 \times \frac{h}{16} \times \frac{w}{16}$.

**Segmentation:** The branch of the network dedicated to segmentation is composed of a single $1 \times 1$ convolution layer (and ReLU non-linearity) followed by a classification layer. The classification layer consists of $h \times w$ pixel classifiers, each responsible for indicating whether a given pixel belongs to the object in the center of the patch. Note that each pixel classifier in the output plane must be able to utilize information contained in the *entire* feature map, and thus have a complete view of the object. This is critical because unlike in semantic segmentation, our network must output a mask for a single object even when multiple objects are present (*e.g.*, see the elephants in Fig. 1).

For the classification layer one could use either locally or fully connected pixel classifiers. Both options have drawbacks: in the former each classifier has only a partial view of the object while in the latter the classifiers have a massive number of redundant parameters. Instead, we opt to decompose the classification layer into two linear layers with no non-linearity in between. This can be viewed as a 'low-rank' variant of using fully connected linear classifiers. Such an approach massively reduces the number of network parameters while allowing each pixel classifier to leverage information from the entire feature map. Its effectiveness is shown in the experiments. Finally, to further reduce model capacity, we set the output of the classification layer to be $h^o \times w^o$ with $h^o < h$ and $w^o < w$ and upsample the output to $h \times w$ to match the input dimensions.

**Scoring:** The second branch of the network is dedicated to predicting if an image patch satisfies constraints (i) and (ii): that is if an object is centered in the patch and at the appropriate scale. It is composed of a $2 \times 2$ max-pooling layer, followed by two fully connected (plus ReLU non-linearity) layers. The final output is a single 'objectness' score indicating the presence of an object in the center of the input patch (and at the appropriate scale).

### 3.2 Joint Learning

Given an input patch $x_k \in \mathcal{I}$, the model is trained to jointly infer a pixel-wise segmentation mask and an object score. The loss function is a sum of binary logistic regression losses, one for each location of the segmentation network and one for the object score, over all training triplets $(x_k, m_k, y_k)$:

$$\mathcal{L}(\theta) = \sum_k \left( \frac{1+y_k}{2w^o h^o} \sum_{ij} \log(1 + e^{-m_k^{ij} f_{segm}^{ij}(x_k)}) + \lambda \log(1 + e^{-y_k f_{score}(x_k)}) \right) \quad (1)$$

Here $\theta$ is the set of parameters, $f_{segm}^{ij}(x_k)$ is the prediction of the segmentation network at location $(i, j)$, and $f_{score}(x_k)$ is the predicted object score. We alternate between backpropagating through the segmentation branch and scoring branch (and set $\lambda = \frac{1}{32}$). For the scoring branch, the data is sampled such that the model is trained with an equal number of positive and negative samples.

Note that the factor multiplying the first term of Equation 1 implies that we only backpropagate the error over the segmentation branch if $y_k = 1$. An alternative would be to train the segmentation branch using negatives as well (setting $m_k^{ij} = 0$ for all pixels if $y_k = 0$). However, we found that training with positives only was critical for generalizing beyond the object categories seen during training and for achieving high object recall. This way, during inference the network attempts to generate a segmentation mask at *every* patch, even if no known object is present.

### 3.3 Full Scene Inference

During full image inference, we apply the model densely at multiple locations and scales. This is necessary so that for each object in the image we test at least one patch that fully contains the object (roughly centered and at the appropriate scale), satisfying the two assumptions made during training. This procedure gives a segmentation mask and object score at each image location. Figure 2 illustrates the segmentation output when the model is applied densely to an image at a single scale.

The full image inference procedure is efficient since all computations can be computed convolutionally. The VGG features can be computed densely in a fraction of a second given a typical input image. For the segmentation branch, the last fully connected layer can be computed via convolutions applied to the VGG features. The scores are likewise computed by convolutions on the VGG features followed by two $1 \times 1$ convolutional layers. Exact runtimes are given in §4.

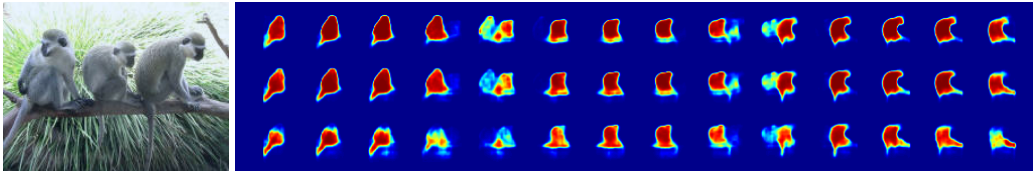

Figure 2: Output of segmentation model applied densely to a full image with a 16 pixel stride (at a single scale at the central horizontal image region). Multiple locations give rise to good masks for each of the three monkeys (scores not shown). Note that no monkeys appeared in our training set.

Finally, note that the scoring branch of the network has a downsampling factor $2\times$ larger than the segmentation branch due to the additional max-pooling layer. Given an input test image of size $h^t \times w^t$, the segmentation and object network generate outputs of dimension $\frac{h^t}{16} \times \frac{w^t}{16}$ and $\frac{h^t}{32} \times \frac{w^t}{32}$, respectively. In order to achieve a one-to-one mapping between the mask prediction and object score, we apply the interleaving trick right before the last max-pooling layer for the scoring branch to double its output resolution (we use exactly the implementation described in [26]).

### 3.4 Implementation Details

During training, an input patch $x_k$ is considered to contain a 'canonical' positive example if an object is precisely centered in the patch and has maximal dimension equal to exactly 128 pixels. However, having some tolerance in the position of an object within a patch is critical as during full image inference most objects will be observed slightly offset from their canonical position. Therefore, during training, we randomly jitter each 'canonical' positive example to increase the robustness of our model. Specifically, we consider translation shift (of $\pm 16$ pixels), scale deformation (of $2^{\pm 1/4}$), and also horizontal flip. In all cases we apply the same transformation to both the image patch $x_k$ and the ground truth mask $m_k$ and assign the example a positive label $y_k = 1$. Negative examples ($y_k = -1$) are any patches at least $\pm 32$ pixels or $2^{\pm 1}$ in scale from any canonical positive example.

During full image inference we apply the model densely at multiple locations (with a stride of 16 pixels) and scales (scales $2^{-2}$ to $2^1$ with a step of $2^{1/2}$). This ensures that there is at least one tested image patch that fully contains each object in the image (within the tolerances used during training).

As in the original VGG-A network [27], our model is fed with RGB input patches of dimension $3 \times 224 \times 224$. Since we removed the fifth pooling layer, the common branch outputs a feature map of dimensions $512 \times 14 \times 14$. The score branch of our network is composed of $2 \times 2$ max pooling followed by two fully connected layers (with 512 and 1024 hidden units, respectively). Both of these layers are followed by ReLU non-linearity and a dropout [28] procedure with a rate of 0.5. A final linear layer then generates the object score.

The segmentation branch begins with a single $1 \times 1$ convolutional layer with 512 units. This feature map is then fully connected to a low dimensional output of size 512, which is further fully connected to each pixel classifier to generate an output of dimension $56 \times 56$. As discussed, there is no non-linearity between these two layers. In total, our model contains around 75M parameters.

A final bilinear upsampling layer is added to transform the $56 \times 56$ output prediction to the full $224 \times 224$ resolution of the ground-truth (directly predicting the full resolution output would have been much slower). We opted for a non-trainable layer as we observed that a trainable one simply learned to bilinearly upsample. Alternatively, we tried downsampling the ground-truth instead of upsampling the network output; however, we found that doing so slightly reduced accuracy.

Design architecture and hyper-parameters were chosen using a subset of the MS COCO validation data [21] (non-overlapping with the data we used for evaluation). We considered a learning rate of .001. We trained our model using stochastic gradient descent with a batch size of 32 examples, momentum of .9, and weight decay of .00005. Aside from the pre-trained VGG features, weights are initialized randomly from a uniform distribution. Our model takes around 5 days to train on a Nvidia Tesla K40m. To binarize predicted masks we simply threshold the continuous output (using a threshold of .1 for PASCAL and .2 for COCO). All the experiments were conducted using Torch7[1].

---

[1] http://torch.ch

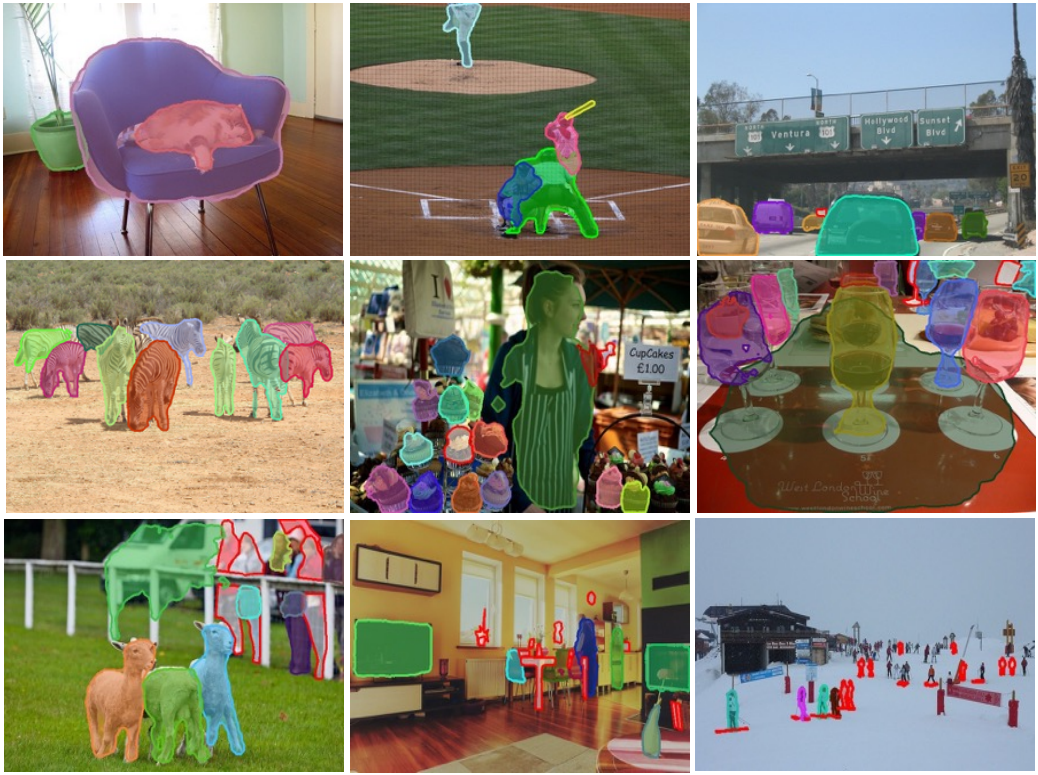

Figure 3: DeepMask proposals with highest IoU to the ground truth on selected images from COCO. Missed objects (no matching proposals with IoU > 0.5) are marked with a red outline.

## 4 Experimental Results

In this section, we evaluate the performance of our approach on the PASCAL VOC 2007 test set [7] and on the first 5000 images of the MS COCO 2014 validation set [21]. Our model is trained on the COCO training set which contains about 80,000 images and a total of nearly 500,000 segmented objects. Although our model is trained to generate segmentation proposals, it can also be used to provide box proposals by taking the bounding boxes enclosing the segmentation masks. Figures 3 and 6 show examples of generated proposals with highest IoU to the ground truth on COCO.

**Metrics**: We measure accuracy using the common Intersection over Union (IoU) metric. IoU is the intersection of a candidate proposal and ground-truth annotation divided by the area of their union. This metric can be applied to both segmentation and box proposals. Following Hosang *et al.* [13], we evaluate the performance of the proposal methods considering the average recall (AR) between IoU 0.5 and 1.0 for a fixed number of proposals. AR has been shown to correlate extremely well with detector performance (recall at a single IoU threshold is far less predictive) [13].

**Methods**: We compare to the current top-five publicly-available proposal methods including: EdgeBoxes [34], SelectiveSearch [31], Geodesic [16], Rigor [14], and MCG [24]. These methods achieve top results on object detection (when coupled with R-CNNs [10]) and also obtain the best AR [13].

**Results**: Figure 4 (a-c) compares the performance of our approach, DeepMask, to existing proposal methods on PASCAL (using boxes) and COCO (using both boxes and segmentations). Shown is the AR of each method as a function of the number of generated proposals. Under all scenarios DeepMask (and its variants) achieves substantially better AR for all numbers of proposals considered. AR at selected proposal counts and averaged across all counts (AUC) is reported in Tables 1 and 2 for COCO and PASCAL, respectively. Notably, DeepMask achieves an *order of magnitude reduction* in the number of proposals necessary to reach a given AR under most scenarios. For example, with 100 segmentation proposals DeepMask achieves an AR of .245 on COCO while competing methods require nearly 1000 segmentation proposals to achieve similar AR.

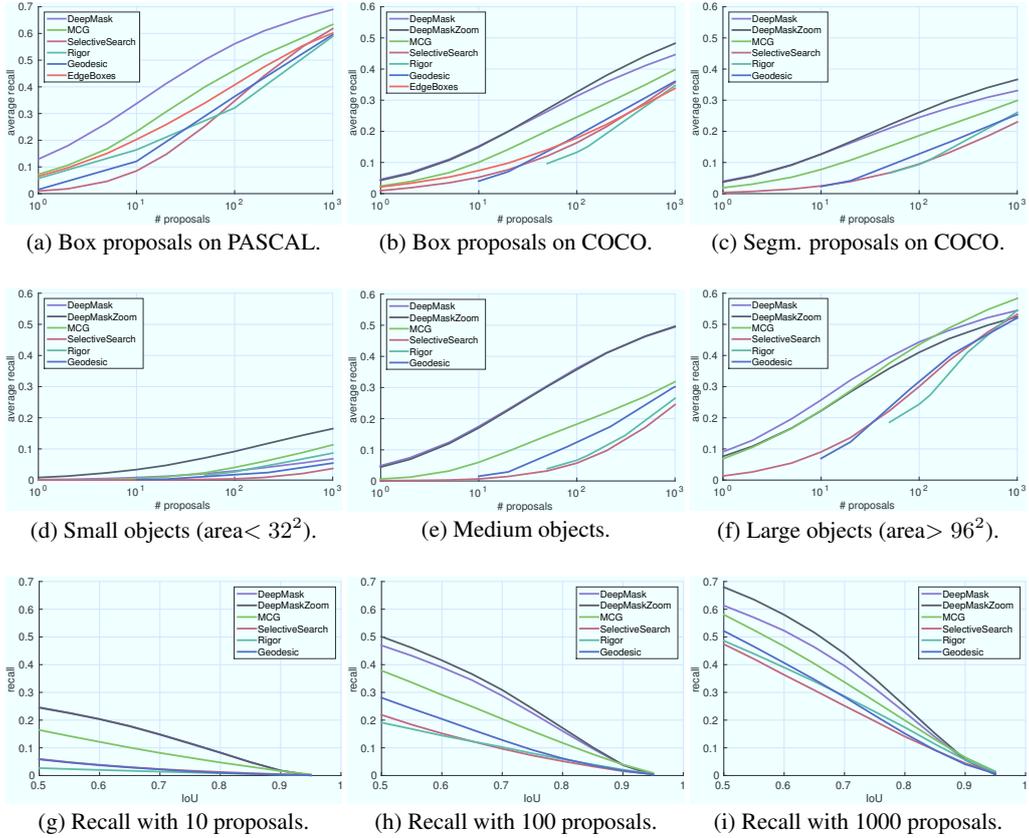

Figure 4: (a-c) Average recall versus number of box and segmentation proposals on various datasets. (d-f) AR versus number of proposals for different object scales on segmentation proposals in COCO. (g-h) Recall versus IoU threshold for different number of segmentation proposals in COCO.

|  | Box Proposals | | | | Segmentation Proposals | | | | | | |
|---|---|---|---|---|---|---|---|---|---|---|---|
|  | AR@10 | AR@100 | AR@1000 | AUC | AR@10 | AR@100 | AR@1000 | AUC$^S$ | AUC$^M$ | AUC$^L$ | AUC |
| EdgeBoxes [34] | .074 | .178 | .338 | .139 | - | - | - | - | - | - | - |
| Geodesic [16] | .040 | .180 | .359 | .126 | .023 | .123 | .253 | .013 | .086 | .205 | .085 |
| Rigor [14] | - | .133 | .337 | .101 | - | .094 | .253 | .022 | .060 | .178 | .074 |
| SelectiveSearch [31] | .052 | .163 | .357 | .126 | .025 | .095 | .230 | .006 | .055 | .214 | .074 |
| MCG [24] | .101 | .246 | .398 | .180 | .077 | .186 | .299 | .031 | .129 | .324 | .137 |
| DeepMask20 | .139 | .286 | .431 | .217 | .109 | .215 | .314 | .020 | .227 | .317 | .164 |
| DeepMask20* | .152 | .306 | .432 | .228 | .123 | .233 | .314 | .020 | .257 | .321 | .175 |
| DeepMaskZoom | .150 | **.326** | **.482** | **.242** | **.127** | **.261** | **.366** | **.068** | .263 | .308 | **.194** |
| DeepMaskFull | .149 | .310 | .442 | .231 | .118 | .235 | .323 | .020 | .244 | **.342** | .176 |
| DeepMask | **.153** | .313 | .446 | .233 | .126 | .245 | .331 | .023 | **.266** | .336 | .183 |

Table 1: Results on the MS COCO dataset for both bounding box and segmentation proposals. We report AR at different number of proposals (10, 100 and 1000) and also AUC (AR averaged across all proposal counts). For segmentation proposals we report overall AUC and also AUC at different scales (small/medium/large objects indicated by superscripts S/M/L). See text for details.

**Scale**: The COCO dataset contains objects in a wide range of scales. In order to analyze performance in more detail, we divided the objects in the validation set into roughly equally sized sets according to object pixel area $a$: small ($a < 32^2$), medium ($32^2 \leq a \leq 96^2$), and large ($a > 96^2$) objects. Figure 4 (d-f) shows performance at each scale; all models perform poorly on small objects. To improve accuracy of DeepMask we apply it at an additional smaller scale (DeepMaskZoom). This boosts performance (especially for small objects) but at a cost of increased inference time.

| PASCAL VOC07 | AR@10 | AR@100 | AR@1000 | AUC |
|---|---|---|---|---|
| EdgeBoxes [34] | .203 | .407 | .601 | .309 |
| Geodesic [16] | .121 | .364 | .596 | .230 |
| Rigor [14] | .164 | .321 | .589 | .239 |
| SelectiveSearch [31] | .085 | .347 | .618 | .241 |
| MCG [24] | .232 | .462 | .634 | .344 |
| DeepMask | **.337** | **.561** | **.690** | **.433** |

Table 2: Results on PASCAL VOC 2007 test.

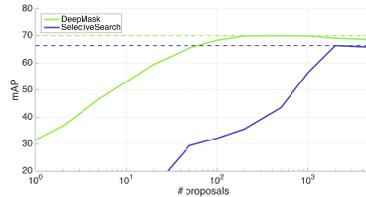

Figure 5: Fast R-CNN results on PASCAL.

**Localization**: Figure 4 (g-i) shows the recall each model achieves as the IoU varies, shown for different number of proposals per image. DeepMask achieves a higher recall in virtually every scenario, except at very high IoU, in which it falls slightly below other models. This is likely due to the fact that our method outputs a downsampled version of the mask at each location and scale; a multiscale approach or skip connections could improve localization at very high IoU.

**Generalization**: To see if our approach can generalize to unseen classes [2, 19], we train two additional versions of our model, DeepMask20 and DeepMask20*. DeepMask20 is trained *only* with objects belonging to one of the 20 PASCAL categories (subset of the full 80 COCO categories). DeepMask20* is similar, except we use the scoring network from the original DeepMask. Results for the two models when evaluated on all 80 COCO categories (as in all other experiments) are shown in Table 1. Compared to DeepMask, DeepMask20 exhibits a drop in AR (but still outperforms all previous methods). DeepMask20*, however, matches the performance of DeepMask. This surprising result demonstrates that the drop in accuracy is due to the discriminatively trained scoring branch (DeepMask20 is inadvertently trained to assign low scores to the other 60 categories); the segmentation branch generalizes extremely well even when trained on a reduced set of categories.

**Architecture**: In the segmentation branch, the convolutional features are fully connected to a $512$ 'low-rank' layer which is in turn connected to the $56 \times 56$ output (with no intermediate non-linearity), see §3. We also experimented with a 'full-rank' architecture (DeepMaskFull) with over 300M parameters where each of the $56 \times 56$ outputs was directly connected to the convolutional features. As can be seen in Table 1, DeepMaskFull is slightly inferior to our final model (and much slower).

**Detection**: As a final validation, we evaluate how DeepMask performs when coupled with an object detector on PASCAL VOC 2007 test. We re-train and evaluate the state-of-the-art Fast R-CNN [9] using proposals generated by SelectiveSearch [31] and our method. Figure 5 shows the mean average precision (mAP) for Fast R-CNN with varying number of proposals. Most notably, with just 100 DeepMask proposals Fast R-CNN achieves mAP of 68.2% and outperforms the best results obtained with 2000 SelectiveSearch proposals (mAP of 66.9%). We emphasize that with $20\times$ *fewer* proposals DeepMask outperforms SelectiveSearch (this is consistent with the AR numbers in Table 1). With 500 DeepMask proposals, Fast R-CNN improves to 69.9% mAP, after which performance begins to degrade (a similar effect was observed in [9]).

**Speed**: Inference takes an average of 1.6s per image in the COCO dataset (1.2s on the smaller PASCAL images). Our runtime is competitive with the fastest segmentation proposal methods (Geodesic [16] runs at ~1s per PASCAL image) and substantially faster than most (*e.g.*, MCG [24] takes ~30s). Inference time can further be dropped by ~30% by parallelizing all scales in a single batch (eliminating GPU overhead). We do, however, require use of a GPU for efficient inference.

## 5 Conclusion

In this paper, we propose an innovative framework to generate segmentation object proposals directly from image pixels. At test time, the model is applied densely over the entire image at multiple scales and generates a set of ranked segmentation proposals. We show that learning features for object proposal generation is not only feasible but effective. Our approach surpasses the previous state of the art by a large margin in both box and segmentation proposal generation. In future work, we plan on coupling our proposal method more closely with state-of-the-art detection approaches.

**Acknowledgements:** We would like to thank Ahmad Humayun and Tsung-Yi Lin for help with generating experimental results, Andrew Tulloch, Omry Yadan and Alexey Spiridonov for help with computational infrastructure, and Rob Fergus, Yuandong Tian and Soumith Chintala for valuable discussions.

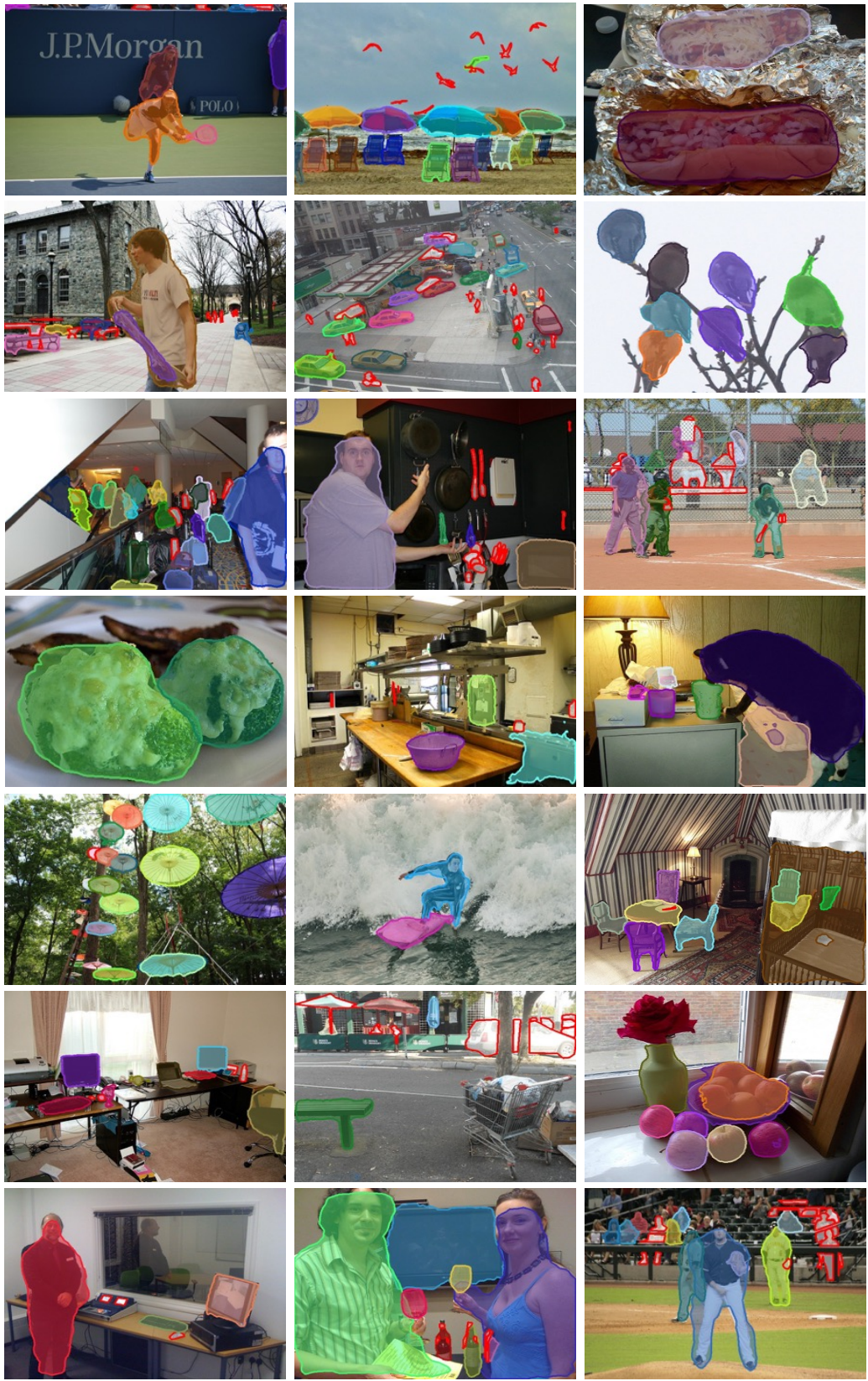

Figure 6: Additional DeepMask proposals with highest IoU to the ground truth on selected images from COCO. Missed objects (no matching proposals with IoU > 0.5) are marked with a red outline.